# DECENTRALIZED DECISION MAKING AND NAVIGATION STRATEGY FOR TRACKING INTRUDERS IN A CLUTTERED AREA BY A GROUP OF MOBILE ROBOTS

by

Muhammad Usman Arif

18 May 2020

i

# TABLE OF CONTENTS

# ABSTRACT

In the current era of industrial revolution, the mobile robots are playing pivotal role in helping out mankind in many complex and hazardous environments for performing tasks like search and rescue, obstacle avoidance, mining and security surveillance etc. A lot of navigation algorithms has been developed in recent years but novel challenges still exist in autonomous path planning of multiple robots to track and follow multiple intruders. This report demonstrates a decentralized strategy of arithmetic mean based navigation algorithm for a group of mobile robots to navigate through an unknown environment filled with obstacles to detect and follow multiple invading intruders. The suggested navigation strategy ensures that mobile robots safely move right in the middle of surrounding obstacles to maintain safe distance and to avoid collision with obstacles and each other. The conventional method of color recognition is used to detect dynamic intruders and calculate pixel values using Microsoft Kinect sensor camera. A probability of danger algorithm is introduced to ensure that all the intruders present in the environment are being followed by friendly robots on the bases of minimum distance between intruder and its follower. The mobile robots follow intruders movement on the bases of their pixel values. The low pixel value means that intruder is far away and high pixel value represents that intruder is closer to the friendly robots. All the algorithms and image processing technique are implemented and tested in WEBOTS simulation environment using C programming language and the results show the success of proposed arithmetic mean based navigation and probability of danger based intruders following algorithms.

# CHAPTER 1.    INTRODUCTION

The development of industrial automation applications in manufacturing, military operations, healthcare, oil and gas, agriculture and mining etc. has resulted in use of mobile robots for performing multiple tasks in an industrial environment [1]. The ground based mobile robots can be used in many complex and hazardous environments for performing tasks like search and rescue, obstacle avoidance, path planning, relocating objects, navigation control, intrusion detection, and object trapping etc. The basic problems of autonomous ground robots involve decentralized navigation and detection of targets. So development of an algorithm, for reliable navigation control and path planning for multiple robots while avoiding obstacles present in their path and finding the multiple target intruders, is of great practical importance and a hot topic of discussion among researchers in last decade.

The navigation control system of a robot to work autonomously in real time involves three functions which are path planning through obstacles [2], self-localization in suggested environment [3], and motion control to reach the final goal [4]. First of all, the part of path planning involves the decision making of a robot to find an optimal way to avoid all the possible obstacles present in the environment. Secondly, the robot should be able to locate its exact position in the environment which is dependent on the application for which the robot is being used. Thirdly, robot motion is an important task which makes the robot to perform specific duties according to the application for which it is being used.

There are many types of applications for which ground robots are being used to search and to detect the intruders in different environments [5]. Due to increase in number of hostilities among state actors and non-state actors has resulted in internal security issues of many countries around the world [6]. These security issues has not only resulted in loss of precious lives of armed forces protecting the border but also putting the lives of common citizens of a country at risk. So the use of ground robots in these sensitive and critical areas around the border for surveillance and detecting intruders will be of great benefit, not only for protecting the precious lives of armed forces but also less human physical involvement will be required. The robots used for these type of applications should be equipped with sensors which enable them to explore the environment to detect the obstacles and intruders present. These surveillance type robots will require enhanced capabilities of navigation and vision to avoid obstacles and detect intruders.

A lot of research papers which will be discussed in Chapter 2 of literature review will demonstrate that a lot of work has been done for navigation of ground robots in different environments. Firstly, many researchers such as [7], [8], [9], [10], [11], [12]; focused on developing different algorithms for robot navigation for avoiding obstacles. But in these research papers all the results were shown in 2D simulation environment. So there is still some area of research which is required to show the results in 3D simulation environment. Secondly, most of the researchers such as [7], [8], [10], [12], [13], [14]; focused on navigation of single robot in an environment. So there is still an area of research for developing strategies for navigation of multiple robots in same environment or experimental terrain. Thirdly, many researchers focused on developing strategies for a robot navigation to reach a fixed target position. Although research is also available for finding targets in unknown position but still research in this area is in early stages and more effective strategies can be developed. Lastly, following moving targets or intruders is not an old problem and currently a lot of research and strategies are being proposed such as [13], [15] and [16] etc. But still there is work needs to be done for detecting multiple intruders present in an area and a method which enable friendly robots to trap or follow all the intruders present.

## 1.1 Problem Statement

In recent years, service based robots became popular and started being used for many application such as vacuum cleaners, food delivery and smart wheelchair, which only required the functionality of avoiding obstacles and mostly fed with the final target positions. But search based robots are different than service based robots. Search based robots require some additional functionalities which provides novel challenges like vision capability to detect specific targets and a navigation algorithm which enable them to operate autonomously irrespective of other friendly robots present in the environment. The navigation strategy needs to be developed for robots to pass through obstacles so that they maintain a safe distance from all the obstacles. The vision capability is required to recognize moving hostile targets or intruders present in the area of interest. These intruders can be at any location, so the robots should be able to explore the environment in such a way that they are able to locate the intruders and then an algorithm is required which will enable them to follow the intruder which is closer to them. An image processing technique will be required which will equip the friendly robots to recognize an intruder in their field of view. A strategy is required which will make sure that friendly robots should not collide with each other and maintain a safe distance between themselves too.

## 1.2 Research Questions

Here are some of the challenging research questions which are very important for understanding the purpose of this report

- What type of robot should be used which will provide a control system feasible for integrating all the sensors and equipment required for the completion of this report?
- What type of algorithm should be developed which will enable all the friendly robots to navigate through obstacles and also enable them to operate autonomously in a decentralized fashion while avoiding collision with each other?
- What type of vision technology and image processing technique should be used to identify the intruder in a specific environment and what type of strategy should be implemented which will enable friendly robots to differentiate among different intruders?
- What type of algorithm should be devised which will enable a friendly robot to follow a specific intruder robot if multiple intruder robots present in their field of view?
- What type of software and simulation environment should be used to show the final results and limitations faced in the report?

**1.3　Aims and Objectives**

The need for developing a decentralized navigation system for multiple robots operating in a cluttered area and detecting intruders has resulted in motivation for doing research and devising a strategy to overcome the proposed challenges. So following are the aims and objectives of this research report

- To develop a strategy for 'decentralized multiple robots path planning' while moving through cluttered environment
- To construct a method for detecting intruders whose positions are unknown
- To develop a mechanism for trapping or following intruders
- To develop a control system for robots while using a camera, range sensor and distance sensors

**1.4　Methodology**

The development of strategies for robot navigation and detection of intruders in an environment filled with obstacles divided into three parts. First of all, an arithmetic mean distance based algorithm is developed for navigation of robot between obstacles. This arithmetic mean based navigation algorithm uses onboard sonar sensors of robot to measure the distance from obstacles on right and left hand side continuously and calculate the average of values between both sides to determine the next points where robot should go and navigate easily between obstacles without colliding. Secondly, the robot will be able to detect the intruders differentiated by different colors in the environment using a camera installed above robot and an image processing algorithm. Thirdly, a range sensor will be used to measure the distance of all the intruders from the robot. The range sensor will also be installed above the robot. A probability of danger algorithm is devised which enable a robot to follow the intruder which is closer than other intruders. The detail of these algorithms and strategies will be discussed in Chapter 3 of this report. The complete algorithm is programmed using C language and same program will be used for all the controllers of the friendly robots. All the algorithm strategies are tested separately and then together as a complete control program.

## 1.5 Structure of the Report

The remainder of this research report discusses the background and literature review in Chapter two which validate the aims and objectives described above. The detailed methodology adopted, algorithms and programming is presented in Chapter three. Chapter four discusses the results obtained via simulation on the software and discussion of obtained results in relation to aims. Chapter five concludes and discusses the finding of this report and presents the future possibilities of research.

# CHAPTER 2.     LITERATURE REVIEW

In this chapter, a careful literature review of navigation strategies of a robot avoiding obstacles and finding target using vision sensors is done. The three most important parts of this research report are robot navigation through obstacles, detecting an intruder using vision sensor and following or trapping the intruder.

## 2.1  Robot Navigation

There are many algorithms proposed by researchers over the years for navigation of ground robots avoiding obstacles and reaching a target point. Here are some of the important works presented by authors

In [7], the author in his PhD dissertation discussed the robot navigation bug algorithm in 2 dimensional environment and introduced tangent bug algorithm for finding shortest path to a final set point. Although robot was operating in an unknown environment for finding shortest path but still final position needed to be set for termination of robot program. Although significant improvement in results were recorded but still there were many shortcomings which needed attention. First of all, bug algorithm performance differs significantly depending on the environment it is being operated. Secondly, all the tests were conducted in 2 dimensional environment. Thirdly, moving around an obstacle region and hope to find a final position may not be fully autonomous system.

In [10] and [17], the authors discussed the development of A star (A*) algorithm for collision free path planning for a robot. A star algorithm is very popular technique for determining the minimum distance between two points. This algorithm detects the corners of the obstacles present in robots field of view and make a pre-emptive decision to follow a path which will avoid obstacles. The results showed the increased efficiency in determining the shortest distance with decreased time. But the issue with this type of technique is the size of the environment. If the size of environment increased then execution speed of algorithm will be decreased. Also this algorithm is developed for only one robot operating in an environment and does not discuss group of robots working in the same environment.

In [18], [19], [20]; the researchers presented a fuzzy logic controller based algorithm for navigating a robot in the environment. A fixed target point is achieved by moving a robot from a starting point. Infrared sensors installed on the robot which detected the obstacles around the robot and give this information to controller which apply fuzzy logic algorithm and change the speeds of right and left wheels of the robot. This approach does introduce a new algorithm for navigating through not only static obstacles but also dynamic obstacles. Although this algorithm is suitable for dynamic obstacles detection but still it does not address the issue of dynamic target. So this approach is only applicable when targets final position is known.

In [9], the author discussed the navigation control of a team of robots moving in unknown environment with obstacles and to trap a target while maintaining a minimum distance among themselves. This work highlights the smart use of energy consumption and complex computations at fixed intervals. All the robots are autonomous in making decision for choosing the best path available. The algorithm developed is focused on reducing the computational time and consumption of energy. Significant results were recorded but in this research work all the robots were fed with information of final target which is static.

In [21], the author introduced a new hybrid algorithm of navigation for robot moving in an environment with dynamic obstacles and goals. This algorithm uses a mechanism by detecting obstacles and goals using a camera. The encouraging results were obtained which were better than earlier algorithms proposed for navigation of robot through dynamic obstacles. But still this algorithm will not able perform if multiple robots are employed in same environment for various tasks.

In [22], the author discussed the algorithm development for a robot navigation in an unknown environment for finding objects and relocating those objects to a specific place. For navigation and finding an exploration path in an unknown environment bug algorithm was used. Secondly, for object detection, which is required to be relocated, image processing techniques like color segmentation and template matching were used by installing a camera above robot. Thirdly, two types of robots were used to measure the efficiency and performance of different robotic control systems for performing the same task. Again a significant improvement of results in 3 dimensional environment were recorded. But there are still many research issues which could be included for improving performance. First of all, image processing techniques used require complex computation capacity and more processing time. So this has made one of the robot slower. Secondly, as robot is moving along the boundary following bug algorithm and searching for objects and starting point at the same point. If the robot find starting point first instead of target object then robot will consider the search process is fully complete and all other targets are not reachable. Also the targets used for experimentation are considered static instead of dynamic. Thirdly, two robots were designed of different structure and tested separately instead of group of robots operating together in same environment. One robot performed better due to its designing and construction.

In [23] and [24], the author introduced a position estimation switching algorithm (PESA) for avoiding convex obstacle and switching navigation algorithm to avoid multi-obstacles in a region by a group of robots in a 2D area. The team leader of robots decides the path to follow in few simple computations and other robots follow the leader. The results does validate the efficacy of proposed algorithms but scope only limited to obstacle detection. In the same way, the authors proposed a distributed control strategy for detection of intruder entering in a protected area by group of robots in [25]. Another paper [26], discussed the probability of intruder detection by a group of robots using a intelligent game theoretic approach. In these two papers of intruder detection, the simulation was done in 2D environment without the deployment of obstacles in the region. Lastly, another algorithm [27], which was formulated to avoid dynamic obstacles with characteristics of changing shapes and sizes. This algorithm considered obstacles as virtual repulsive force and region between obstacles as virtual attractive force to navigate a robot through the region. Again scope is limited to avoid obstacles in the region and simulation is conducted in 2D environment. So there is still an open area of research, in which a fleet of robots should be able to detect and avoid obstacles while keeping track of intruders in an environment.

## 2.2 Target detection using vision sensor

The availability of low cost, low power, high resolution and high speed processing cameras has made it very commonly used for vision purpose in ground robots these days [28]. The purpose of camera is to provide data, composed of images of the environment, to the robot. The robot then apply some image processing technique to derive some desired information which is used for detecting an intruder and also useful for making the robot to follow the intruder. Following are some of the vision sensors and techniques being used by researchers

In [14], the author designed a robot using a camera which detects the target object by using image segmentation technique and also follow the target object maintaining a specific distance. The technique used is of great importance as it is also being used to great extent in current report. But there are some areas where further research could have been done. Only image processing technique is used in this report and no algorithm proposed in case of any obstacle present in the path of robot. The experimental results shown by this experiments used only one target object.

The most relevant research report using a camera above robot for stereo vision and convolutional neural network tracker algorithm to follow a person is presented by [15]. The robot able to detect a moving person using camera in such a way that enable it to calculate the distance from target person. The convolutional neural network algorithm enable the robot to recognize the direction of target person in which it is moving. So even when the target is not in the field of view of robot, it will still be able to follow the target depending on its last known location. This algorithm showed significant results following a moving target. But still there are some areas of research which are ignored in this report which include only one moving target is considered during experimentation and no obstacle avoidance algorithm used.

In the following research papers [5],[13], [16], [29], [30]; the authors used Microsoft Kinect sensor to detect target objects. The Kinect sensor has an RGB camera and rangefinder which enables the robot to not only detect the target object but also the distance from the target object. This Kinect sensor is very useful for calculating the distance of robot from multiple and follow them. The drawbacks of these papers include only one target object or static target objects.

In [22], the author described many image processing techniques such as image segmentation [31], template matching [32], scale invariant feature transform (SIFT) [33] and speeded up robust features (SURF) [34]. Firstly, the image segmentation is very simple and basic technique which detects different colors and textures in an image. So if intruders present in an environment be given a specific color, then it will be very easy to detect them by using simple image segmentation technique. Secondly, in template matching a robot is fed with the template of object image which needed to be detected. If template of object image matches in image provided by vision sensor then its very easy to detect intruders. The template of object image will always be smaller than actual template of image received from camera. Thirdly, SIFT and SURF are some advanced image processing techniques which can be used to detect even fine features of an object in an image. As high computation is required for template matching, SIFT and SURF image processing techniques, so simple image segmentation is very suitable for this research report.

## 2.3 Conclusion

So researchers have devised many algorithms and techniques for robot navigation avoiding obstacles and detecting and following targets using vision sensors. Almost all the algorithms have their own limitations and better results can be achieved by combining different algorithms and techniques together. First of all, for navigation of robot through obstacles no already proposed algorithm like Bug, A star, Fuzzy logic or new hybrid is used for this report. Instead a simple algorithm based on arithmetic mean values of distances from obstacles is formulated which enables the robot to navigate through obstacles without colliding them. Secondly, intruders are detected by using color segmentation technique as compared to other image processing techniques as this requires less computation and processing time. Thirdly, instead of using camera and distance sensor separately, Kinect sensor is used as it provides both camera for vision and rangefinder to measure the distance from intruders. An algorithm of probability of danger is developed which provide the friendly robots to follow the target which is nearer to them. The detail of all these algorithms and techniques will be described in next Chapter three.

The novelty of this research report is going to be the originality of probability of danger and arithmetic mean based algorithms which are specifically not been described in any research paper described above. Also the simplicity of sensors interaction with each other and achieving the goal by using limited number of sensors is also an important aspect of this research report.

# CHAPTER 3.     METHODOLOGY AND ALGORITHMS

## 3.1   Introduction

An environment or terrain, in which a robot move, usually have a starting point from where the robot will start its movement, a final target or goal which the robot has to achieve and multiple obstacles of different sizes and shapes. The aim of robot is to move continuously for finding an optimal path that enables it to achieve final goal while avoiding obstacles in its path. The system which enables the robot to achieve all these tasks is known as navigation system. The navigation system of robot uses the essential data or information of the environment which it gathers by using onboard sensors or devices for deciding its movement. As the final target or intruder position is unknown so robot should plan its movement in such a way that enables it to detect target at any position of the environment [22]. The environment considered for this research report is an indoor environment that means it is surrounded by boundary walls and robot will have to perform all the tasks within these walls.

In this chapter all the designing tools and software use for this project are discussed. The algorithm which is being used for navigation of robot in designed environment is elaborated in detail. The use of image sensor and image segmentation technique, which are used to detect the intruders, are discussed. After that the probability of danger algorithm is discussed which makes the robot intelligent enough to follow the nearest intruder. All the limitations and assumptions which are considered in designing the environment and execution of program in this report are also discussed separately. In the last part, the programming or code which is used is explained.

## 3.2   Designing Tools

### *3.2.1   Software and Programming Language used*

First of all, there many software and programming tools which are being used to program robots by researchers like ROS (robot operating system), MATLAB, CYBERBOTICS and GAZEBOISM etc. all over the world. The software which is used for programming and designing of simulation environment for this report is WEBOTS [35]. WEBOTS is an open source robot simulator. This simulator is commonly being used around the world in different universities for research purpose. The advantages of using this software are that it provides the ability to design 3D environment which is just same as a robot operating in real environment and it also supports multiple programming language such as C, C++, Python, MATLAB and java etc. Due to advantage of using multiple programming language in a single software make it easy to do complex tasks like image processing.

All the robots controllers are programmed using basic C language. Basically a single program algorithm is developed which is being operated on all the controllers of the robots. The are some built-in libraries in WEBOTS software and some basic libraries of C language which are used for programming of robots.

*3.2.2    Robot and Sensors Selection*

The robot selected for programming and simulation for this report is Pioneer 3-DX which is part of Pioneer family of robots [36]. This robot device is being used around the world by many advanced research institutions. As designing a robot is not a part of this project, so selecting Pioneer 3-DX as a test robot has provided all the necessary onboard tools and sensors which is enough for execution of all the tasks intended as shown in Figure 1.

The Pioneer 3-DX has 16 onboard ultrasonic (sonar) sensors from which are 8 sensors are on front side and 8 on rear side. All the sensors are equally spaced covering 20 degrees of angle between each other except for four side sensors which have angle of 40 degrees as shown in Figure 2 [37].

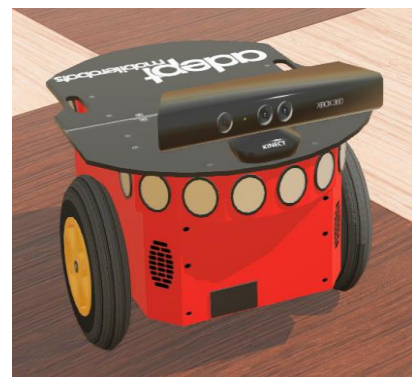

**Figure 1: Pioneer 3-DX with Kinect Sensor**

For this particular report only front side eight sonar sensors will be used to measure the distance from obstacles as there is no need to measure the distance on rear side of the robots. Sonar sensors operate in different way than conventional distance measuring sensors, which means if reflected rays from obstacles are greater than the angle of 45 degrees of reflection cone of aperture never return the distance value as explained in Figure 3 [38]. The sonar sensors can detect an obstacle within 5 meters distance.

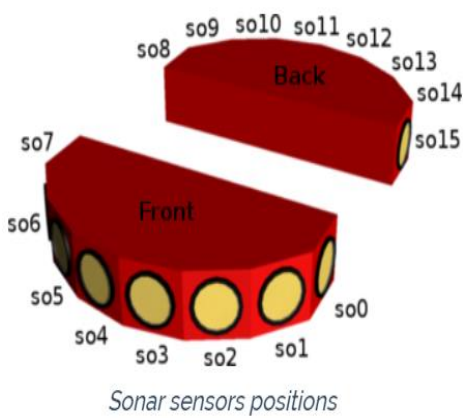
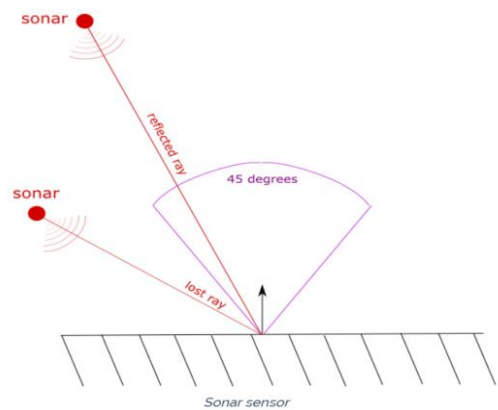

**Figure 2: Sonar sensors position**

**Figure 3: Reflection of sonar ray**

### *3.2.3 Kinect Sensor for vision and depth measurement*

Now for detection of intruders in the simulation environment, a vision sensor (camera) is required. Apart from vision, a distance sensor or range finder is also required which will measure the distance of robot from the intruders. Both of these functionality can be achieved by connecting a simple Microsoft Kinect sensor with a robot as shown in Figure 1. The Kinect sensor, which is shown in Figure 4 [39], is easily available in market and has been used by many researchers in their projects as already explained in chapter two.

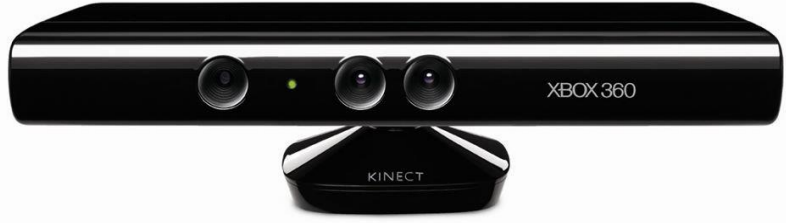

**Figure 4: Kinect sensor with camera and range finder capability**

### 3.3 Navigation Path Algorithm

There are many algorithms for path planning of a robot avoiding obstacles such as A star, Bug and fuzzy logic controller algorithms etc. which has been discussed in chapter two. The most popular among these algorithms is Bug algorithm in which a robot follow the wall of obstacle till it reaches the final goal position. But in this report, any algorithms discussed above, has not been used. But a new strategy of measuring distances from obstacles on left and right hand side of a robot and taking their arithmetic mean value to decide the navigation of robot in the simulation environment is used.

#### *3.3.1 Arithmetic Mean Based Navigation Strategy*

As ultrasonic (sonar) sensors are installed on Pioneer robot, they will measure the distance from obstacles and navigate the robot through obstacles. So a force or velocity will be required to move the robot wheels. Hence

$$V = \{v_{1(t)}, v_{2(t)}, \ldots \ldots, v_{n(t)}\} \qquad (1)$$

$$W = \{w_{1(t)}, w_{2(t)}, \ldots \ldots, w_{n(t)}\} \qquad (2)$$

Where $V$ is the linear velocity of a robot and $W$ is the angular velocity of the robot at any time t and n is the number of robots present in the simulation environment. Now at any instant during simulation the linear and angular velocity should always satisfy the following condition

$$Vmin < v_{1(t)} = v_{2(t)} = \cdots = v_{n(t)} < Vmax \qquad (3)$$

$$Wmin < w_{1(t)} = w_{2(t)} = \cdots = w_{n(t)} < Wmax \tag{4}$$

Now let's consider $So_{(n)}$ be the ultrasonic (sonar) sensors installed on Pioneer 3-DX robot where n is the total number of sonar sensors. As there are total 8 sensors installed on robot so n = (0,1,2,3,….,7). Now four sensors will be present on left front side and four sensors on right front side of the robot. So let's consider

The left front side sensors of the robot will be donated as

$$So_m^L \quad \text{where} \quad \forall\, m \epsilon n,\, m = 0,1,2,3 \tag{5}$$

So all the sensors on left side can be denoted in an array

$$So_m^L = [So_0^L, So_1^L, So_2^L, So_3^L] \tag{6}$$

The right front side sensors of the robot will be donated as

$$So_k^R \quad \text{where} \quad \forall\, k \epsilon n,\, k = 4,5,6,7 \tag{7}$$

So all the sensors on right side can also be represented in an array

$$So_k^R = [So_4^R, So_5^R, So_6^R, So_7^R] \tag{8}$$

We suppose that a robot, represented as $R_1$, is passing through an environment filled with obstacles. Now when a robot navigate between left and right side obstacles then it will detect the distance from them using sonar sensors in such a way as shown in Figure 5. As it has already been mentioned that only those sonar sensors rays will be detectable which are within their detectable range. So

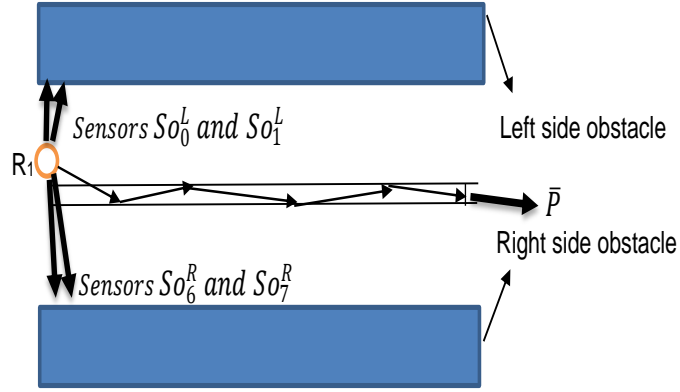

**Figure 5: Robot movement between obstacles**

In the above scenario, two sensors on left side has detected the obstacle and two sensors on right side has detected the obstacle. Then equation (6) and (8) will become like $So_m^L = [So_0^L, So_1^L, 0, 0]$ and $So_k^R = [0, 0, So_6^R, So_7^R]$ respectively. So all the values of sensors which are zero will be ignored and rest of the values will be saved. As a first step of arithmetic mean based algorithm let's use the values of left side sonar sensors of robot

$$\overline{So_m^L} = \sum_{P=0}^{3} So_P^L \Big/ Total\ no.\ of\ nonzero\ values, where\ So_P^L > 0 \quad (9)$$

If same mechanism is applied to right side values then

$$\overline{So_k^R} = \sum_{P=4}^{7} So_P^R \Big/ Total\ no.\ of\ nonzero\ values, where\ So_k^L > 0 \quad (10)$$

Now after calculating the mean values of distances on each side of robot let's calculate the arithmetic mean value of these obtained values.

$$\bar{P} = \overline{So_m^L} + \overline{So_k^R} \Big/ 2 \quad (11)$$

Hence, $\bar{P}$ is the point where the robot needs to go next as shown in Figure 5. So theoretically, $\bar{P}$ will be point which is approximately in the middle of obstacles.

Now to navigate the robot to move either towards left or right side following conditions are applied

- If $\bar{P} < \overline{So_m^L}$ then robot right wheel velocity will decrease and left wheel velocity will remain same. This will make the robot to move towards right side until following condition is not met

$$-0.015 < \bar{P} - \overline{So_m^L} < 0.015 \qquad (12)$$

If following condition satisfied then velocity of both wheels will become same again and robot will try to stay within threshold limit as defined in above equation. This minimum threshold limit is defined after experimenting different values of threshold in the simulation environment.

- If $\bar{P} > \overline{So_m^L}$ then robot left wheel velocity will increase and right wheel velocity will remain same. This will make the robot to move towards left side until following condition is not met

$$-0.015 < \bar{P} - \overline{So_k^R} < 0.015 \qquad (13)$$

If following condition is also satisfied then again robot will try to stay within threshold limit.

So the above algorithm or strategy for robot navigation does ensure that robot safely pass through area filled with obstacles without any danger of collision as robot sensors are continuously measuring the distance from obstacles and updating the next point where robot should go. Furthermore, the strength and weaknesses of this proposed strategy will be discussed in Chapter 5.

**3.4  Intruder Detection via Image Processing**

Now after developing a mechanism for robots navigation system to avoid obstacles, the second task is to detect intruders in the simulation environment. As these intruders are dynamic and can change their location, so a vision sensor is going to be required. For that the camera of Kinect sensor is going to be used as already mentioned earlier in the designing tools section. To detect intruders in the simulation environment, image processing technique will be required. There are many image processing techniques like template matching, color segmentation and speeded up robust features etc. But for this project color segmentation technique is applied for detecting intruders.

### *3.4.1 Intruder Detection by Colour Segmentation*

Color segmentation is one of the simple and basic image processing technique which is being used by researchers to detect different colored objects in an image. Basically all the colors are made by mixing the different concentrations of red, green and blue known which are also known as RGB. As an image is comprised of pixels and each pixel can be processed in such a way that intensity of each color of RGB can be measured. The intensity value for each color varies between 0 to 255 and can be saved in 1 byte (8 bits). So if intensity of R is 255, G is 0 and B is 0 in a pixel of image then that image is of red color. As values of RGB is being calculated for each pixel of a image, so returned data is of 24 bits.

As an example if color segmentation is applied on an image for detecting red color as shown in Figure 6 which is taken from [22] in such a way that detected red pixels should be represented as white pixels and all other pixels should be black then as a result we get image as shown in Figure 7 [22].

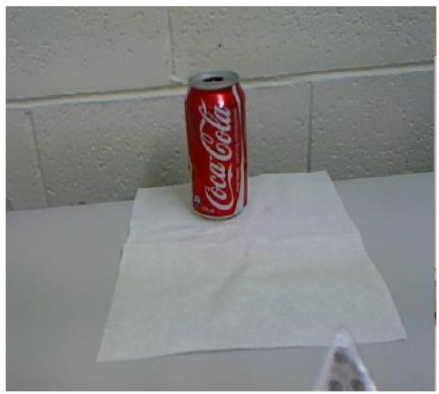 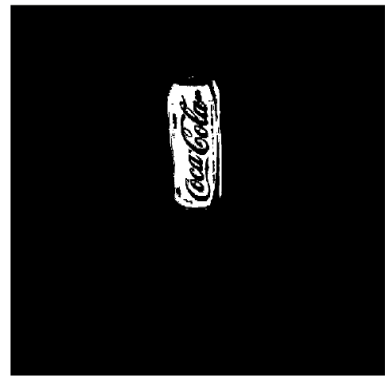

**Figure 6: Camera image of can    Figure 7: Color segmented image**

In this project report, two intruder robots needs to be detected as shown in Figure 8. Each intruder robot is assigned by a different color so that searching robot can differentiate between them. When image processing technique is applied to detect them via camera, the result is shown in Figure 9.

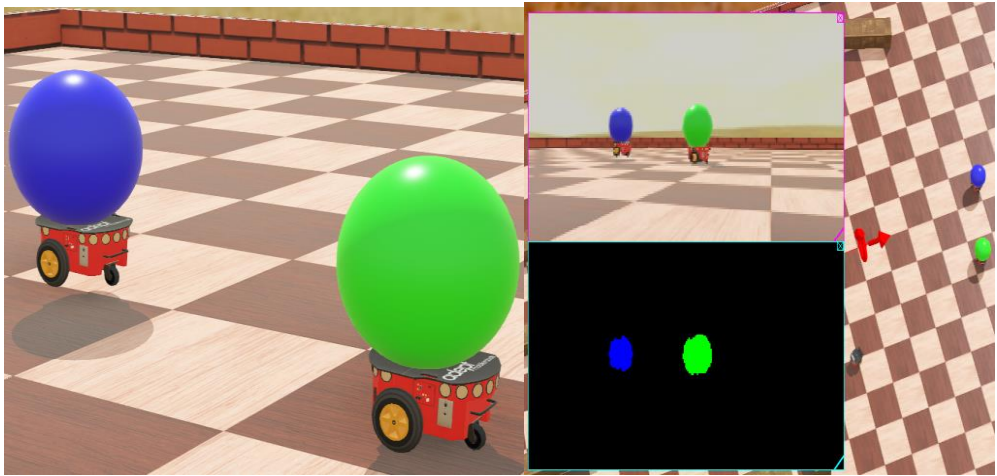

**Figure 8: Intruders**        **Figure 9 :Color segmented image**

As successful results for detecting are achieved by just applying simple image processing technique so there is no need to opt for complex image processing techniques. As this software simulation environment does behave same like real world environment, so even if light intensity changes in the environment the robot will still be able to detect the intruders based on their colors.

**3.5   Probability of Danger Algorithm**

After successful navigation between obstacles and detection of intruders by a robot, the next task is to either follow the intruder if it is moving or trap it if it is stationary. The friendly robots should maintain a specific distance from the hostile intruders. The friendly robots should also be able to avoid collision with each other. After assigning colors to intruders, it is very easy for friendly robots to ignore all other robots in the environment and consider them as obstacles. So when a friendly robot come in range of another robot, then same algorithm of arithmetic mean to avoid obstacles will apply.

To follow only one intruder is very easy for a group of robots when they detect it by using camera. On the other hand, there are two intruders being deployed in the simulation environment for this project. So friendly group of robots has to make a decision to follow intruders in such a way that each intruder is being followed by at least one robot. So, to achieve this task a probability of danger algorithm is devised.

Generally, consider if there are two hostile targets or intruders which are approaching towards a person then the target which is closer to the person is more dangerous than other. Same technique is also applied to make a decision by robot to follow an intruder in this report. So the robot should follow the intruder which is closer to it, if multiple intruders are in its field of view. The friendly robots first detect that which target is nearer than other, then robots should be able to calculate distance from each of them. Hence, a range finder is used which comes as a function or part of Kinect sensor. By using range finder or also known as depth finder robot will calculate the distance from intruders in its field of view as shown in Figure 10

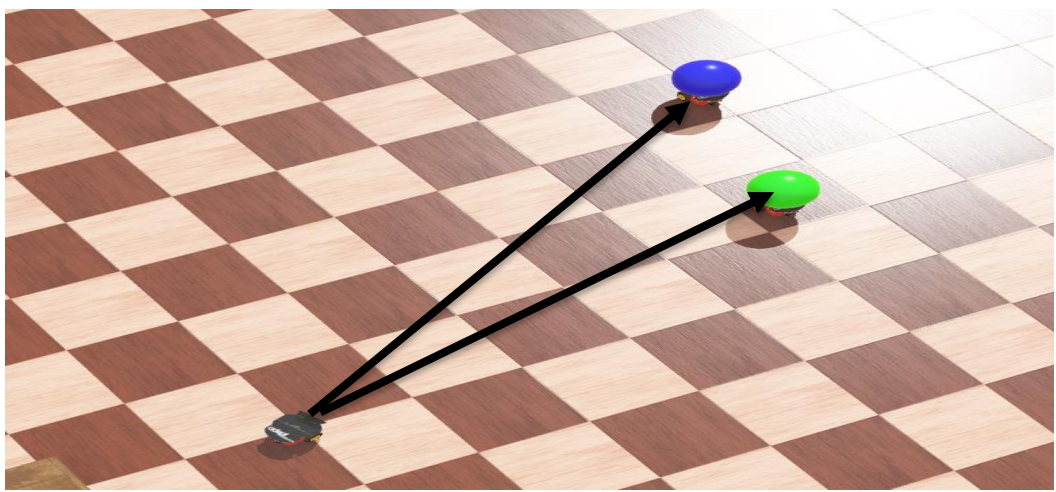

**Figure 10: Distance between intruders and robot**

In above figure, robot should be able to make the decision to follow green robot as it is closer than the blue robot. This algorithm also works in situations when a friendly robot is following an intruder and suddenly another intruder comes in field of view, which is closer than the intruder which robot is following, the friendly robot will change the target and start following the robot which is closer now.

This probability of danger technique for following multiple robots in a simulation environment is very simple and important but novel in its own way. This technique make this project unique and special in this regard.

### 3.6 Assumptions

There are multiple assumptions which are made during the designing of simulation environment for this report and they are as follows

- All the sensors used in this project will return perfect values which is not the case in real world environment where sensors are prone to return false or incorrect values due to noise.
- There is no separate program written for intruders to move freely in the simulation environment. But these intruders can easily be moved manually with the help of keyboard and mouse in the software. But during the simulation in this project, readily available code in the libraries of WEBOTS software for moving intruder in a straight line is used.
- A minimum distance is considered between the obstacles so that robot can easily navigate between obstacles.
- All the mechanical parts and sensors used in this project will work perfectly and no mechanical issue will be faced as all the work is done in simulation environment.
- There are four friendly robots used for navigation between obstacles and intruder detection. Two intruder robots are used which are assigned different colors so that they can be detected easily using camera.
- For applying probability of danger algorithm, the simulation environment is designed in such a way that both intruders will be in the field of view of friendly robots so that each one of them can make a decision by using danger algorithm.

### 3.7 Programming and algorithms

Pseudo-code or algorithm which is followed in implementation of all the proposed techniques is shown in below pseudo-code Figure 11

```
1. Initialize all robots, sensors and devices
2. Check for Obstacles in surrounding area
3. If (obstacles detected) then
       Measure distance from obstacles
       Apply arithmetic mean based navigation algorithm
4. Else If (no obstacle detected) then
       Find intruders using camera
       Apply color segmentation technique
           If (multiple intruders detected) then
               Calculate distance from intruders using range finder
               Apply probability of danger algorithm
               Follow the intruder nearer to the robot
           Else if ( single intruder detected)
               Calculate distance from intruder
               Follow intruder
             If (detected pixels of intruder which robot is following > 3000)
          then
                Robot is in stationary position
             Else
                 Keep following

       If (obstacle is detected again)

       Go to Step 3
```

**Figure 11: Pseudo-code of robot navigation and intruder detection**

The algorithm which is used to detect and follow intruder is given in Figure 12

1. Divide camera image in 2 parts (Left half of image & Right half of image)
2. Do image processing on left half of image (color segmentation)
    A. Detect and count blue intruder pixels in left half
    B. Detect and count green intruder pixels in left half
3. Do image processing on right half of image
    A. Detect and count blue intruder pixels in right half
    B. Detect and count green intruder pixels in right half
4. Calculate distance from each robot
    A. If (range from blue intruder < range from green intruder)
        Add blue robot total pixels and ignore green intruder
    B. Else if(range from blue robot > range from blue robot)
        Add green robot pixels and ignore blue robot
5. For following intruder left and right
    A. If (pixels on right half  < pixels on left half)
        Move robot to left
    B. Else if (pixels on right half > pixels on left half)
        Move robot to right
    C. Else if (pixels on right half ==pixels on left)
        Go straight
6. If (total detected pixels < 10)
    Keep searching Intruder
   Else if ( total detected pixels > 3000)
    Stop Robot

**Figure 12: Intruder detection and following algorithm**

The flowchart of above strategy is shown in below Figure 13

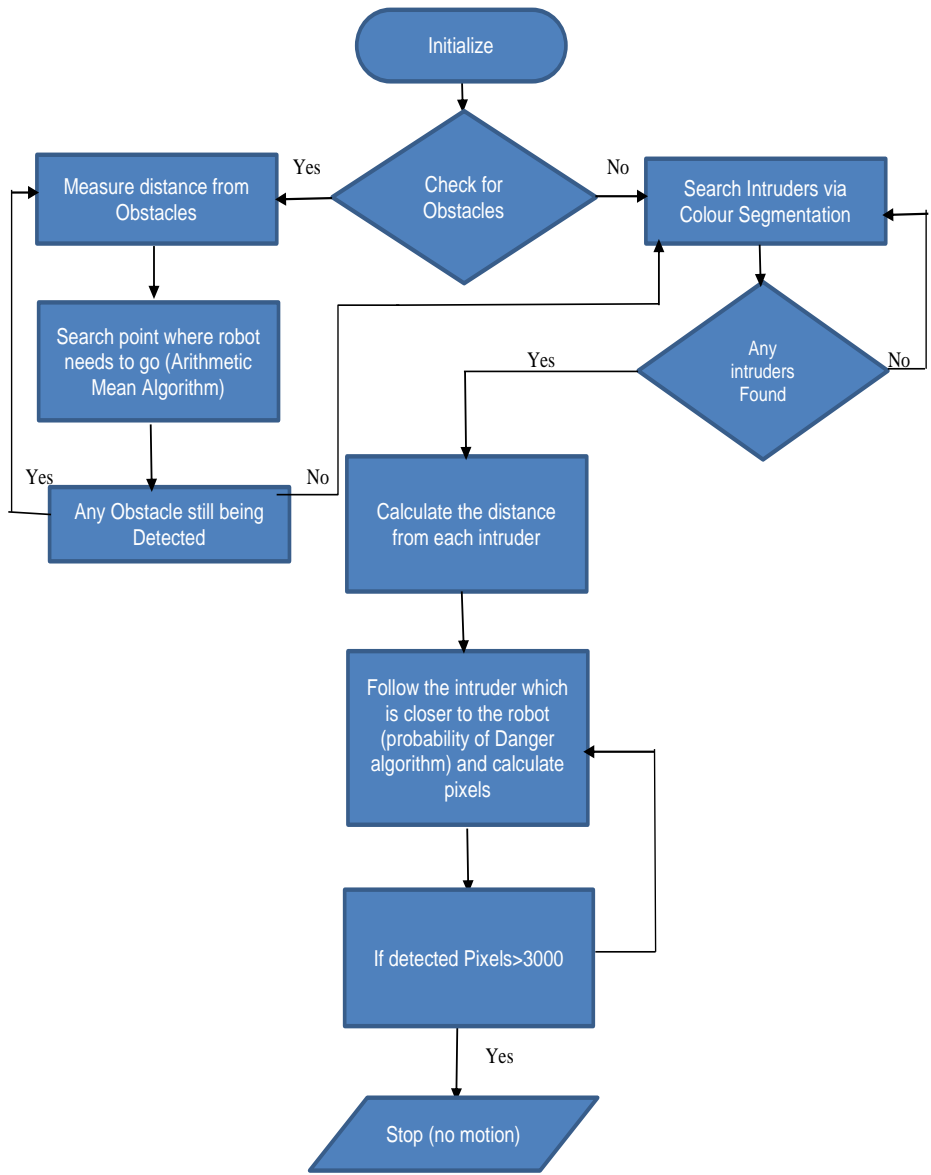

**Figure 13: Flowchart of robot navigation and intruder detection**

As already mentioned that basic C language was used for the implementation of all the algorithms in WEBOTS software. In below Figure 14 all the libraries and basic functions used for programming for this project are displayed.

```c
1 // All the libraries used in this thesis project of robot navigation and intruder detection
2 #include <webots/robot.h>
3 #include <math.h>
4 #include <stdio.h>
5 #include <stdlib.h>
6 #include <string.h>
7 #include <webots/distance_sensor.h>
8 #include <webots/led.h>
9 #include <webots/gps.h>
10 #include <webots/motor.h>
11 #include <webots/camera.h>
12 #include <webots/range_finder.h>
13 #include <webots/display.h>
14 #include <assert.h>
15 #include <time.h>
16 // how many sensors are on the robot and maximum value returned by the sensors
17 #define MAX_SENSOR_NUMBER 8
18 #define MAX_SENSOR_VALUE 1024
19 #define TIME_STEP 64
20 // maximum speed allowed
21 #define MAX_SPEED 5.24
22 #define PI 3.14159265
```

**Figure 14: Libraries used for programming**

After defining the libraries used in this project, below Figure 15 is displaying function in which all the robots, devices and sensors is initialized

```c
23 // For initializing all the robots, devices and sensors used
24 int main(int argc, char **argv) {
25   wb_robot_init();   // initialize robot
26   // definig kinect sensor parameters
27   static WbDeviceTag kinectColor;   static WbDeviceTag kinectRange;   const float *kinect_values;
28   kinectColor = wb_robot_get_device("kinect color");
29   kinectRange = wb_robot_get_device("kinect range");
30   wb_camera_enable(kinectColor, TIME_STEP);
31   wb_range_finder_enable(kinectRange, TIME_STEP);
32   const int kinect_width = wb_range_finder_get_width(kinectRange);
33   const int kinect_height = wb_range_finder_get_height(kinectRange);
34   const double max_range = wb_range_finder_get_max_range(kinectRange);
35   // initialize camera and display features. Display window will be used for showing image after colour segmentation
36   WbDeviceTag camera = wb_robot_get_device("camera");
37   WbDeviceTag camera_display = wb_robot_get_device("camera_display");
38   wb_camera_enable(camera, TIME_STEP);
39   wb_camera_recognition_enable(camera, TIME_STEP);
40   //attach the camera with the display
41   int display_width = wb_display_get_width(camera_display);
42   assert(display_width == wb_camera_get_width(camera));
43   int display_height = wb_display_get_height(camera_display);
44   assert(display_height == wb_camera_get_height(camera));
45   wb_display_attach_camera(camera_display, camera);
46   //for sonar sensors
47   int i,j;
48   WbDeviceTag distance_sensor[8];
49   char distance_sensor_names[8][4] = {
50     "so0", "so1", "so2", "so3",
51     "so4", "so5", "so6", "so7"
52   };
53   // initialize devices
54   for (i = 0; i < 8 ; i++) {
55     distance_sensor[i] = wb_robot_get_device(distance_sensor_names[i]);
56     wb_distance_sensor_enable(distance_sensor[i], TIME_STEP);           }
57   // for intitlaizing robot wheels
58   WbDeviceTag left_wheel = wb_robot_get_device("left wheel");
59   WbDeviceTag right_wheel = wb_robot_get_device("right wheel");
60   wb_motor_set_position(left_wheel, INFINITY);
61   wb_motor_set_position(right_wheel, INFINITY);
62   wb_motor_set_velocity(left_wheel, 0.0);
63   wb_motor_set_velocity(right_wheel, 0.0);
64   int leftobsmatrix[4];    int rightobsmatrix[4];
```

**Figure 15: All devices and sensors initialized**

Now in main function first of all the distance (sonar) sensor will measure whether there is an obstacle is around or not. If there is an obstacle then arithmetic mean based navigation strategy will be implemented as code for that is shown in below Figures 16 and 17 from code line 73 to line 140.

```c
65  // main loop where all the algorithms are implemented
66  while (wb_robot_step(TIME_STEP) != -1) {
67  // activating gps and getting their values
68    double leftsum = 0 ; double m=0 ; double rightsum=0; double n=0 ; double leftaverage = 0 ; double rightaverage = 0;
69    double left_speed  = 0.5 * MAX_SPEED;   double right_speed = 0.5 * MAX_SPEED;     // Maximum speed allowed for robot to move
70  // for reading distance sensor values
71    double distance_sensor_values[8];
72    // for left side obstacles
73    for (j = 0; j < 4 ; j++) {
74    distance_sensor_values[j] = wb_distance_sensor_get_value(distance_sensor[j]);
75    leftobsmatrix[j] = distance_sensor_values[j];  }
76     printf("left side obstacles values:");
77     for(j=0 ; j<4 ;j++) {
78    printf("%d ", leftobsmatrix[j]);
79       if(j==3){
80         printf("\n");}
81       if (leftobsmatrix[j] > 0) {
82    leftsum = leftsum + leftobsmatrix[j];
83    m = m+1; }
84    else if (leftobsmatrix[j] == 0) {
85    m= m;
86    leftsum = leftsum;  }  }
87     if ( m>0 ) {
88    leftaverage = leftsum / m ;
89    }
90      else if ( m ==0 ) {
91    leftaverage = leftsum;
92    }
93    printf("sum of left matrix is: %f ", leftsum);
94    printf("total left values: %f ", m);
95    printf("average of left matrix is: %f ", leftaverage);
96    printf("\n");
97    // Right obstacle matrix
98    for (j = 4; j < 8 ; j++) {
99    distance_sensor_values[j] = wb_distance_sensor_get_value(distance_sensor[j]);
100   rightobsmatrix[0] = distance_sensor_values[4];
101   rightobsmatrix[1] = distance_sensor_values[5];
102   rightobsmatrix[2] = distance_sensor_values[6];
103   rightobsmatrix[3] = distance_sensor_values[7];
104   }
105   // right obstacle matrix
106   printf("right side obstacles values:");
```

Figure 16: Algorithm for navigation (1)

```c
107       for(j=0 ; j<4 ;j++) {
108         printf("%d ", rightobsmatrix[j]);
109         if(j==3){
110           printf("\n"); }
111         if (rightobsmatrix[j] > 0) {
112    rightsum = rightsum + rightobsmatrix[j];
113    n = n+1;}
114    else if (rightobsmatrix[j] == 0) {
115    n= n;
116    rightsum = rightsum;}}
117    if ( n >0 ) {
118    rightaverage = rightsum / n ;   }
119    else if ( n == 0) {   rightaverage = rightsum;   }
120    printf("sum of right matrix is: %f ", rightsum);
121    printf("total right values: %f ", n);
122    printf("average of right matrix is: %f ", rightaverage);
123    printf("\n");
124    double distance_left = 5.0 * (1.0 - (leftaverage / MAX_SENSOR_VALUE));
125    printf("robot to left obs distance is: %f \n", distance_left);
126    double distance_right = 5.0 * (1.0 - (rightaverage / MAX_SENSOR_VALUE));
127    printf("robot to right obs distance is: %f \n", distance_right);
128    double robot_set_value = (distance_left + distance_right)/2;
129    printf("robot should go to : %f  \n", robot_set_value);
130    if (robot_set_value < 3.2 ){
131    left_speed  = 0.5 * MAX_SPEED;
132    right_speed = 0.5 * MAX_SPEED;
133    if (robot_set_value - distance_left   >0.015 ) {
134     left_speed  += 0.5 ;
135     right_speed -= 0.5 ;    }
136    else if (robot_set_value - distance_left  < -0.015 ) {
137    left_speed  -= 0.5 ; }
138    right_speed += 0.5 ; }
139    else { left_speed =0.5* MAX_SPEED;
140          right_speed =0.5* MAX_SPEED;   } }
141    bool find_target =
142    distance_sensor_values[0] < 700.0 &&
143    distance_sensor_values[1] < 700.0 &&
144    distance_sensor_values[2] < 700.0 &&
145    distance_sensor_values[3] < 700.0 &&
146    distance_sensor_values[4] < 700.0 &&
147    distance_sensor_values[5] < 700.0 &&
148    distance_sensor_values[6] < 700.0 &&
149    distance_sensor_values[7] < 700.0; // if no obstacle deteced
```

Figure 17: Algorithm for navigation control (2)

Now when there is no obstacle around the robot then it will try to find the intruder by using color segmentation technique. The robot is programmed in such a way that when it detects an intruder to follow after applying probability of danger algorithm, the robot will try to position itself according to strategy shown in Figure 12 (where algorithm is described in detail), that selected intruder will always be present in the middle of field of view of robot as code shown below Figures 18, 19 and 20 from line 164 to 249.

```
151    if (find_target) {
152    // *************** For camera image and getting RGB values   ***********
153        const unsigned char *image = wb_camera_get_image(camera);
154        int image_width = wb_camera_get_width(camera);
155        int image_half_width = image_width/2;
156        int image_height = wb_camera_get_height(camera);
157         int image_half_height = image_height/2;
158         float total_pixel = 32768;
159         float total_half_pixel = total_pixel/2;
160         int detected_pixel_green_left = 0; int remaining_pixel = 0; int detected_pixel_left = 0; int detected_pixel_right = 0;
161         int detected_pixel_blue_left = 0; int detected_pixel = 0;  int detected_pixel_green_right = 0; int detected_pixel_blue_right = 0;
162          float value; float range_from_blue_robot = 10 ; float range_from_green_robot = 10 ;
163    // ***************** USE OF KINECT SENSOR AND RANGE FINDER VALUES  *******************
164            kinect_values = (float *)wb_range_finder_get_range_image(kinectRange);
165     // ******For Left hand side half image processing********
166        for (int x = 0; x < image_half_width; x++) {
167        for (int y = 0; y < image_height; y++) {
168        float r = wb_camera_image_get_red(image, image_width, x, y);
169        float g = wb_camera_image_get_green(image, image_width, x, y);
170        float b = wb_camera_image_get_blue(image, image_width, x, y);
171        //******* for blue enemy robot
172        if (b < 130 || r > 134 || g > 134)          // for non blue pixels
173        {
174         wb_display_set_color(camera_display, 0xFFFFFF);
175         wb_display_draw_pixel(camera_display, x, y);    }
176        if (b > 130 && r< 134 && g < 134)            // for blue pixels
177        { wb_display_set_color(camera_display, 0x0000FF);
178         wb_display_draw_pixel(camera_display, x, y);
179         detected_pixel_blue_left = detected_pixel_blue_left + 1;
180         value = wb_range_finder_image_get_depth(kinect_values, kinect_width, x, y);
181           if ( value < range_from_blue_robot )   {
182            range_from_blue_robot = value;
183         }     }
184        else  if(r > 60 || b > 0 || g >0 )      {
185         wb_display_set_color(camera_display, 0xFFFFFF);
186         wb_display_draw_pixel(camera_display, x, y);         }
187        //******** for green enemy robot
188        if (b < 116 && r < 116 && g > 161)     // for green pixels
189        {
190        wb_display_set_color(camera_display, 0x00FF00);
191        wb_display_draw_pixel(camera_display, x, y);
192         detected_pixel_green_left = detected_pixel_green_left + 1;
193        value = wb_range_finder_image_get_depth(kinect_values, kinect_width, x, y);
```

**Figure 18: Intruder detection and following code (1)**

```
            detected_pixel_green_left = detected_pixel_green_left + 1;
            value = wb_range_finder_image_get_depth(kinect_values, kinect_width, x, y);
                if ( value < range_from_green_robot )   {
                    range_from_green_robot = value;     }     }
            else   if(r > 60 || b > 0 || g >0 )
            {wb_display_set_color(camera_display, 0xFFFFFF);
             wb_display_draw_pixel(camera_display, x, y);}       }
             printf("detected pixel of blue enemy on left=%d \n ", detected_pixel_blue_left);
             printf("detected pixel of green enemy on left=%d \n ", detected_pixel_green_left);
             //********For right hand side half image processing**********
             for (int a =image_half_width; a < image_width; a++) {
             for (int c = 0 ; c < image_height; c++) {
             float r = wb_camera_image_get_red(image, image_width, a, c);
             float g = wb_camera_image_get_green(image, image_width, a, c);
             float b = wb_camera_image_get_blue(image, image_width, a, c);
             //******* for blue enemy robot
              if (b > 130 && r < 134 && g < 134)  {           // for blue color detection
               wb_display_set_color(camera_display, 0x0000FF);
               wb_display_draw_pixel(camera_display, a, c);
               detected_pixel_blue_right = detected_pixel_blue_right + 1;
              value = wb_range_finder_image_get_depth(kinect_values, kinect_width, a, c);
              if ( value < range_from_blue_robot )
              {
              range_from_blue_robot = value;       }    }
                else   if(r > 60 || b > 0 || g >0 )  {
             wb_display_set_color(camera_display, 0xFFFFFF);
             wb_display_draw_pixel(camera_display, a, c);     }
             //******** for green enemy robot
             if (b < 116 && r < 116 && g > 161)       // for green pixels
             {
             wb_display_set_color(camera_display, 0x00FF00);
             wb_display_draw_pixel(camera_display, a, c);
             detected_pixel_green_right = detected_pixel_green_right + 1;
             value = wb_range_finder_image_get_depth(kinect_values, kinect_width, a, c);
                 if ( value < range_from_green_robot )   {
                   range_from_green_robot = value;  }       }
                else   if(r > 60 || b > 0 || g >0 )
             { wb_display_set_color(camera_display, 0xFFFFFF);
              wb_display_draw_pixel(camera_display, a, c);      }     }     }
             printf("detected pixel of blue enemy on right=%d \n ", detected_pixel_blue_right);
             printf("detected pixel of green enemy on right=%d \n ", detected_pixel_green_right);
             printf("Robot distance from blue robot=%f \n ", range_from_blue_robot);
             printf("Robot distance from green robot=%f \n ", range_from_green_robot);
```

**Figure 19: Intruder detection and following code (2)**

```
         printf("Robot distance from green robot=%f \n ", range_from_green_robot);
         if ( range_from_blue_robot < range_from_green_robot)    {
          detected_pixel_left = detected_pixel_blue_left;
          detected_pixel_right = detected_pixel_blue_right;         }
          else if ( range_from_blue_robot > range_from_green_robot)
          { detected_pixel_left = detected_pixel_green_left;
          detected_pixel_right = detected_pixel_green_right;      }
           printf("detected_pixel_right=%d \n ", detected_pixel_right);
          printf("detected_pixel_left=%d \n ", detected_pixel_left);
           detected_pixel = detected_pixel_left + detected_pixel_right;
           printf("detected_pixel=%d \n ", detected_pixel);
        if ( detected_pixel_right <  detected_pixel_left ) {
        left_speed  -= 0.5 ;
        right_speed += 0.5 ;       }
        if ( detected_pixel_right >  detected_pixel_left ) {
        left_speed  += 0.5 ;
        right_speed -= 0.5 ;      }
        if ( detected_pixel_right ==  detected_pixel_left ) {
            left_speed  = 0.5 * MAX_SPEED;
            right_speed = 0.5 * MAX_SPEED;       }
        if (detected_pixel < 10) {
            left_speed   =0;
            right_speed  =0.5 * MAX_SPEED;   }
        if (detected_pixel > 3000) {
            left_speed   =0;
            right_speed  =0;          } }
   wb_motor_set_velocity(left_wheel, left_speed);
   wb_motor_set_velocity(right_wheel, right_speed);
};
wb_robot_cleanup();
return 0;
}
```

**Figure 20: Intruder detection and following code (3)**

Hence the programming strategy which is displayed above has been implemented during the achievement of desired results for this project.

**3.8   Conclusion**

In this chapter, it has been explained in detail the tools and programs which are used for this report project. The algorithm has also been explained in detail. All the strategies and algorithms applied are not of complex nature but unique in their own implementation, particularly probability of danger algorithm and arithmetic mean based strategies of navigation which are the reasons of novelty of this report work.

# CHAPTER 4. RESULTS

In this chapter, results are going to be derived and discussed in detail after implementing the methodology explained in Chapter 3 in WEBOTS software.

## 4.1 Simulation Environment

The simulation environment for obtaining the results for this report is designed in such a way that four friendly robots will move through passage having obstacles on both sides. There are multiple obstacles which are dispersed in the simulation environment that can be viewed in below Figure 21. All the friendly robots are clearly marked as Robot 1, Robot 2, Robot 3 and Robot 4 respectively and intruders can also be viewed named as Intruder 1 and Intruder 2 in below Figure 21.

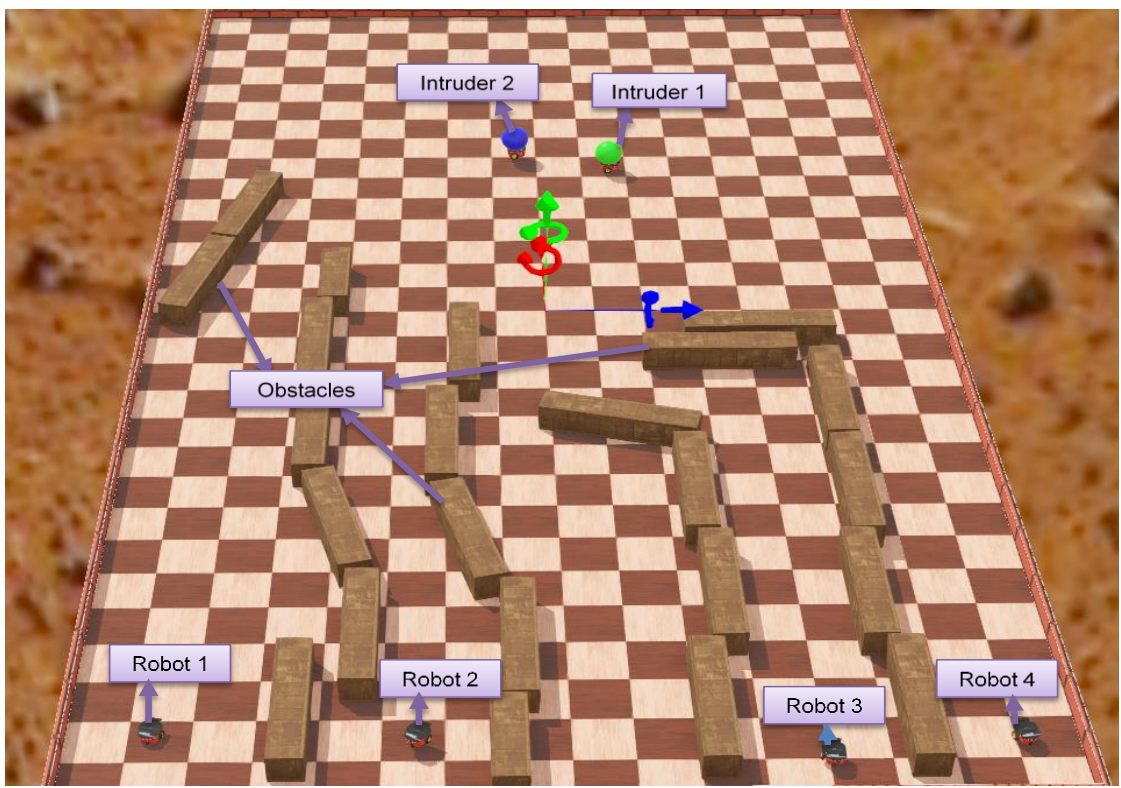

**Figure 21: Design of simulation environment**

The simulation environment is designed intentionally in this way, so that when friendly robots come out of obstacle filled area, they can detect both intruders in their camera field of view which will enable them to apply probability of danger algorithm to decide which intruder they need to follow. In WEBOTS software the required data or results can be viewed in the console which is located at the bottom of software window and can be viewed in below Figure 22.

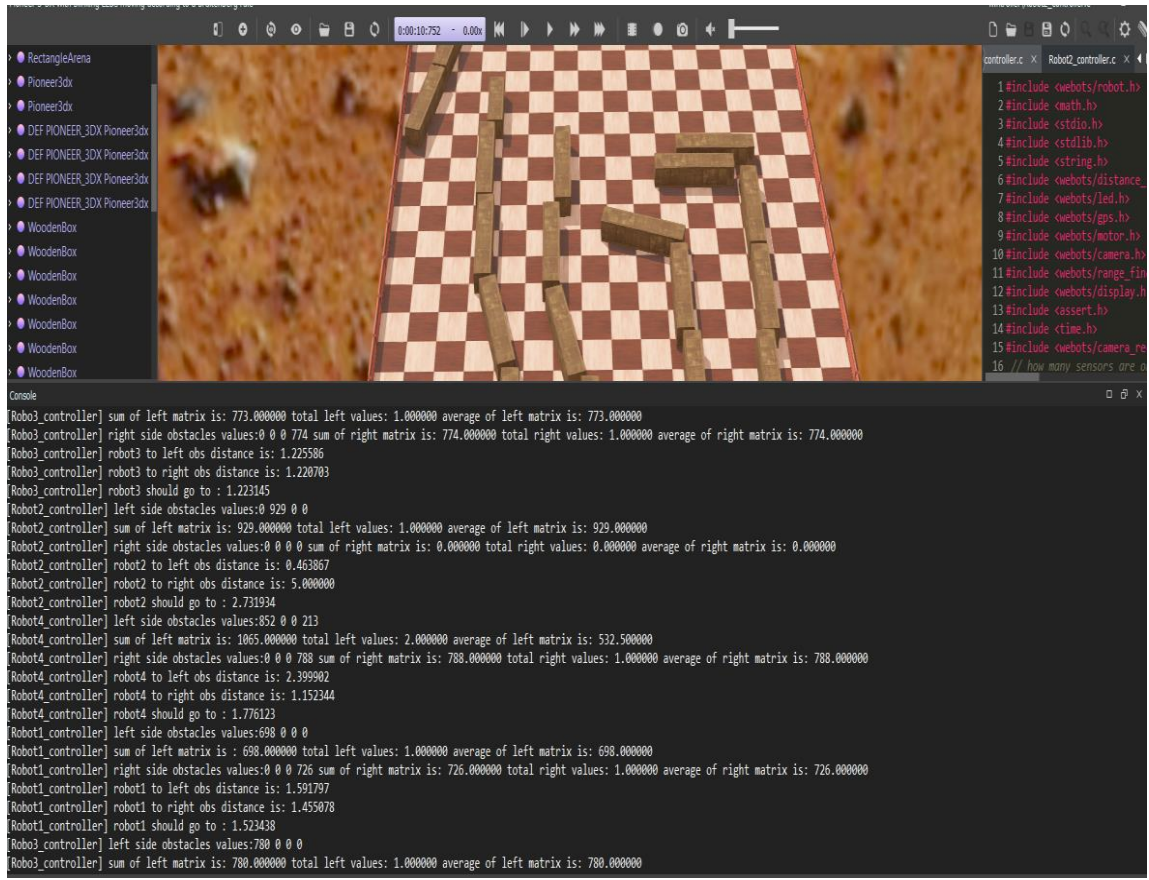

**Figure 22: Data extraction from software**

There are different type of results which needs to be displayed to support the hypothesis, aims and objectives of this project.

## 4.2 Robot Navigation Through Obstacles

First of all, the first objective of this report is the navigation of robot through obstacles which is based on arithmetic mean strategy to avoid obstacles. As already discussed above in Chapter 3 that robot needs to stay approximately in the middle of obstacles during motion so that it can effectively avoid collision.

After implementation of navigation algorithm and running the simulation for 40 seconds, following data is recorded which is shown below in Table 1. All the values are in meters and taken at 12 different intervals of simulation time while robots are passing through obstacles. During this time of simulation the robots are always detecting the obstacles around them as there is no intruder in sight of these robots. There is also one more condition, that when a robot is detecting obstacle only on either left or right side then the side which is not detecting any obstacle within its range of 5 meters will return maximum value of 5 meters. Then arithmetic mean will be taken by considering 5 meter distance on that side.

**Table 1: Data representing distances of robots from obstacle**

| | Time in seconds (s) | 3 | 5 | 8 | 12 | 15 | 19 | 22 | 26 | 29 | 33 | 36 | 40 |
|---|---|---|---|---|---|---|---|---|---|---|---|---|---|
| **Robot 1 Data** | Distance from left side obstacle (m) | 1.59 | 1.56 | 1.5 | 1.62 | 1.66 | 1.59 | 1.55 | 1.51 | 1.59 | 1.43 | 1.54 | 1.49 |
| | Distance from right side obstacle (m) | 1.67 | 1.37 | 1.51 | 2.53 | 2.38 | 2.41 | 2.13 | 1.81 | 1.6 | 1.53 | 1.49 | 1.44 |
| | Total distance between obstacles (m) | 3.26 | 2.93 | 3.01 | 4.15 | 4.04 | 4 | 3.68 | 3.33 | 3.19 | 2.96 | 3.03 | 2.93 |
| | Suggested Position of robot by arithmetic mean algorithm (m) | 1.63 | 1.47 | 1.51 | 2.08 | 2.02 | 2 | 1.84 | 1.66 | 1.59 | 1.48 | 1.51 | 1.46 |
| **Robot 2 Data** | Distance from left side obstacle(m) | 1.79 | 1.47 | 0.43 | 0.81 | 0.96 | 0.84 | 0.64 | 0.7 | 0.76 | 0.94 | 0.73 | 0.78 |
| | Distance from right side obstacle (m) | 1.01 | 1.12 | 1.41 | 1.13 | 0.65 | 0.31 | 0.66 | 0.57 | 0.62 | 0.83 | 0.86 | 0.79 |
| | Total distance between obstacles (m) | 2.79 | 2.6 | 1.85 | 1.94 | 1.61 | 1.16 | 1.3 | 1.26 | 1.38 | 1.78 | 1.59 | 1.57 |
| | Suggested Position of robot by arithmetic mean algorithm (m) | 1.4 | 1.3 | 0.92 | 0.97 | 0.81 | 0.58 | 0.65 | 0.63 | 0.69 | 0.89 | 0.8 | 0.79 |
| **Robot 3 Data** | Distance from left side obstacle (m) | 1.28 | 1.19 | 1.23 | 0.64 | 0.91 | 0.71 | 0.91 | 0.98 | 2.86 | 2.13 | 2.86 | 2.86 |
| | Distance from right side obstacle(m) | 1.23 | 1.13 | 1.22 | 1.04 | 0.81 | 0.86 | 0.83 | 1.06 | 2.69 | 2.36 | 2.23 | 2.23 |
| | Total distance between obstacle (m) | 2.51 | 2.32 | 2.45 | 1.68 | 1.72 | 1.57 | 1.74 | 2.04 | 5.55 | 4.48 | 5.09 | 5.09 |
| | Suggested Position of robot by arithmetic mean algorithm (m) | 1.25 | 1.22 | 1.23 | 0.84 | 0.86 | 0.79 | 0.87 | 1.02 | 2.77 | 2.24 | 2.54 | 2.54 |
| **Robot 4** | Distance from left side obstacle (m) | 0.94 | 0.83 | 0.7 | 1.53 | 1.07 | 1.17 | 0.92 | 1.17 | 3.04 | 0.97 | 0.83 | 0.6 |
| | Distance from right side obstacle (m) | 1.05 | 1.28 | 1.21 | 1.48 | 1.47 | 1 | 1.14 | 1.05 | 1.07 | 1.04 | 1.27 | 1.49 |
| | Total distance between obstacles (m) | 1.99 | 2.1 | 1.91 | 3.01 | 2.54 | 2.16 | 2.06 | 2.22 | 4.12 | 2.01 | 2.09 | 2.09 |
| | Suggested Position of robot by arithmetic mean algorithm (m) | 1 | 0.98 | 0.96 | 1.51 | 1.27 | 1.08 | 1.03 | 1.11 | 2.06 | 1.01 | 1.05 | 1.05 |

Now next step is to make a meaning of this data and represent results in the form of graphs which will represent the effectiveness of arithmetic mean based algorithm for obstacle avoidance.

First of all, let's consider the data for Robot 1 which is detecting obstacles around it. Now when a robot is measuring the distance from left side obstacles and right side obstacles, then it means that total distance between obstacles is the sum of distances on left and right side obstacle measured by the robot. Now let's consider distance of robot from right side obstacle is $D_n^R$ and distance of robot from left side obstacle is $D_n^L$, then total distance between obstacles can be given by $D_n^T$

$$D_n^T = D_n^R + D_n^L \qquad (14)$$

Where n is the robot which is estimating the total distance between obstacles. Now by keeping in view the total distance between obstacles, the navigation algorithm returns a value which will be approximately in the middle of these obstacles where the robot needs to go next.

Now the question to ponder, what is the actual location of robot between the obstacles during navigation?. So exact location of robot between the obstacles is the maximum of distances value of left side obstacle and right side obstacle. Now if distance of robot from left obstacle distance is $D_n^L$ and distance from right obstacle is $D_n^R$ then the robot actual location between obstacles $P_n$ can be represented as

$$P_n = Supremum(D_n^L, D_n^R) \qquad (15)$$

Where n is the robot which is navigating through obstacles. This logic will only apply to the values when obstacle is detected on both sides of the robot. So after using the data from above figure, the path which is being followed by Robot 1 is displayed in below Figure 23

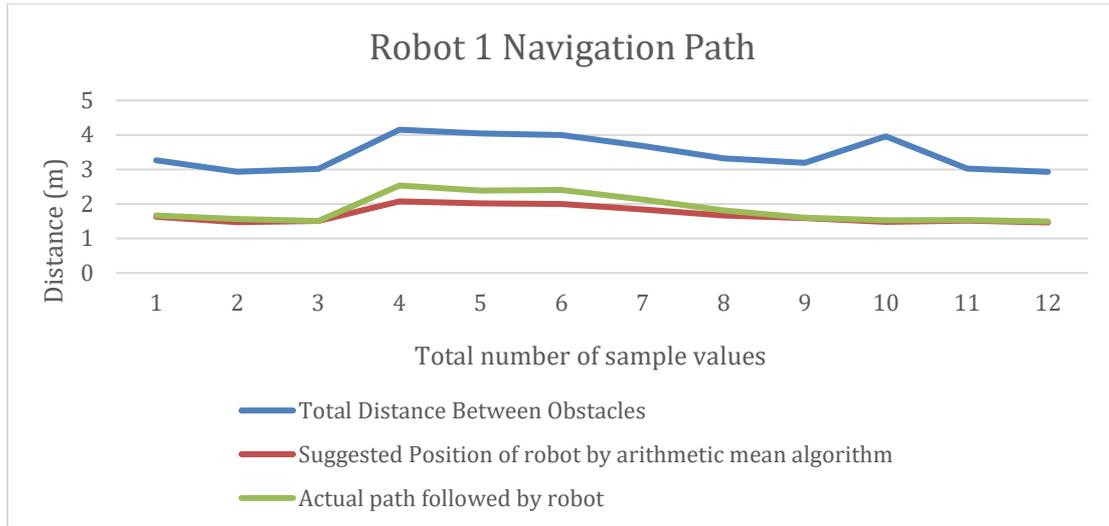

**Figure 23: Robot 1 navigation path**

In the above graph, total distance is of robot 1 is calculated using equation 14 and represented with blue line, while suggested path is calculated using equation 11 and represented with red line and actual path followed by robot is calculated using equation 15 and represented with green line. It can be seen clearly that robot tried to follow the suggested path with minor deviation which is due to slow turn rate of robot due to physical attributes within the simulation environment. If a robot is going in one direction and suddenly an obstacle encountered in the path, then robot will take some time to turn towards new path.

In the same way of designing the Robot 1 navigation path, the navigation path followed by Robot 2 is shown in below Figure 24. Same equations are used to calculate the parameters which are also represented in Table 1

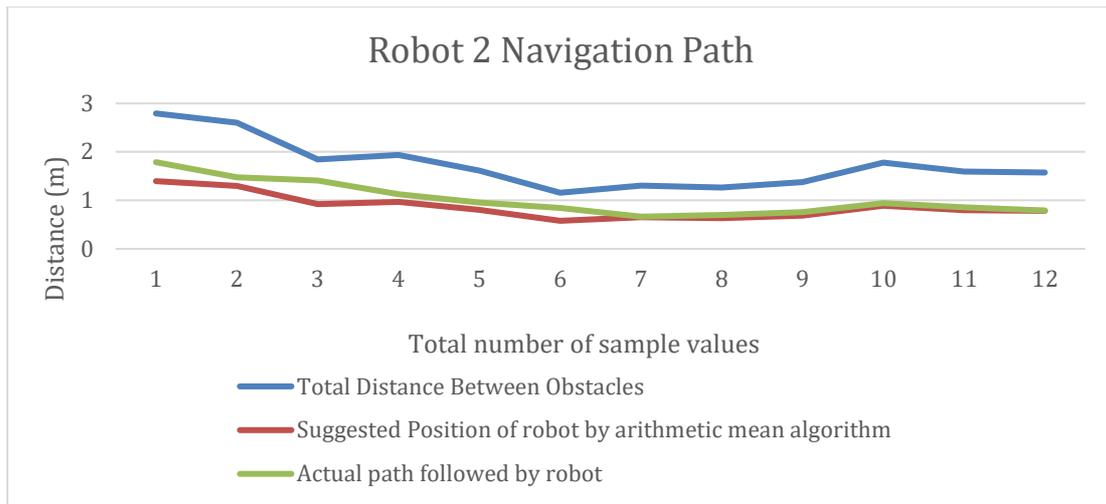

**Figure 24: Robot 2 navigation path**

Again the deviation from suggested path is due to time taken by robot to turn and settle down to new values. But efficacy of suggested algorithm can be viewed clearly that in any instance of time robot was never close to the detected obstacle.

Now the mechanism used for robot 1 and robot 2 will also be applied to represent the results for robot 3 path navigation in Figure 25.

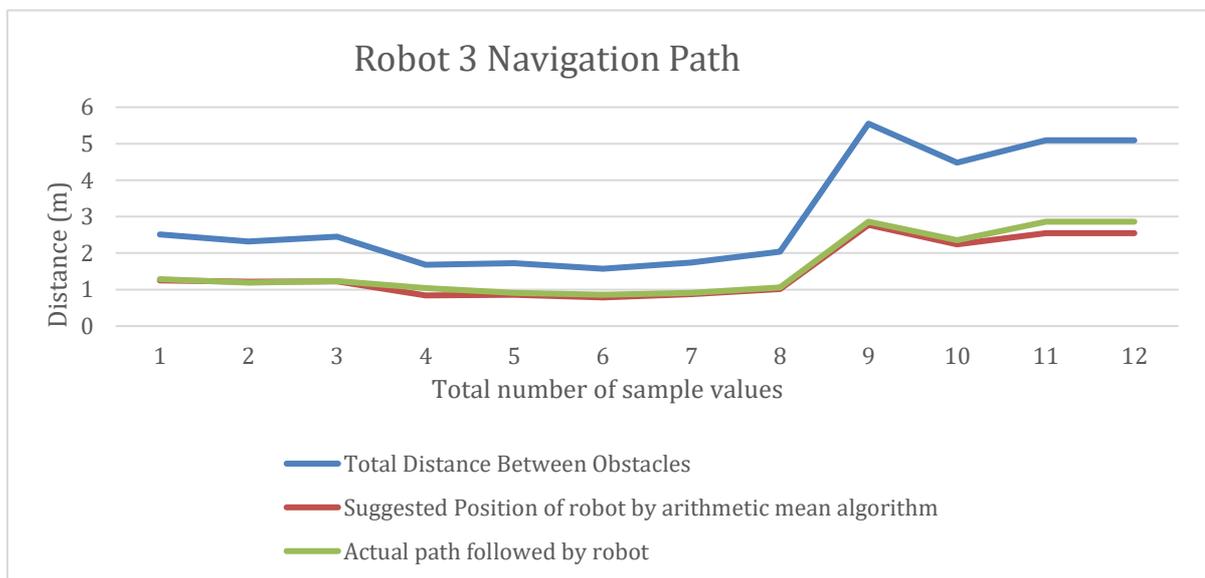

**Figure 25: Robot 3 navigation path**

Robot 3 has actually maintained the best navigation path being followed. This is due to distance between obstacles remained almost constant where robot 3 is navigating.

For robot 4 same strategy and equations are used and results are represented in Figure 26.

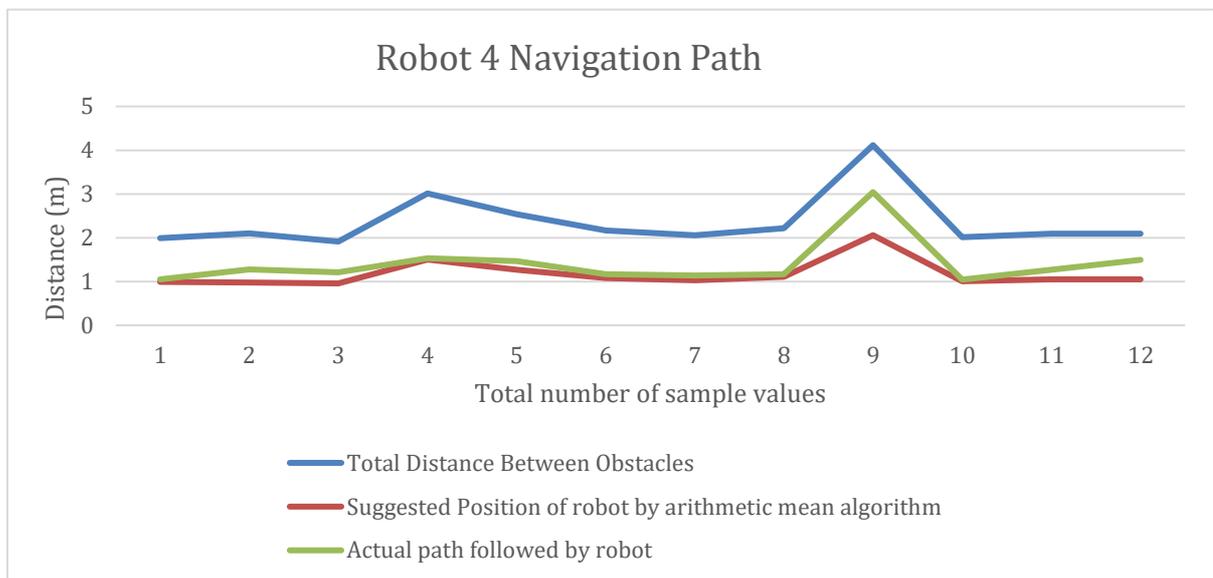

**Figure 26: Robot 4 navigation path**

Again robot 4 has tried to position itself right in the middle of obstacles according to the data collected in Table 1.

Hence, it can be seen clearly from these graphs that arithmetic mean algorithm for navigation of robot is efficient enough to avoid obstacles and keeping the robot right in the middle of obstacles.

**4.3  Intruders Detection and Intruder Following**

For intruder detection, a camera is being used which is part of Kinect sensor which installed above robots as mentioned in Chapter 3. Two types of colors blue and green are assigned to two intruders which are deployed in simulation environment as shown earlier in Figure 21. After intruders are in field of view of camera then robot detects them by using color segmentation technique. As in Figure 27, a color segmented image is shown for Robot 2 in which robot detected blue and green colors. Also shown is the camera view of each robot detecting both intruders.

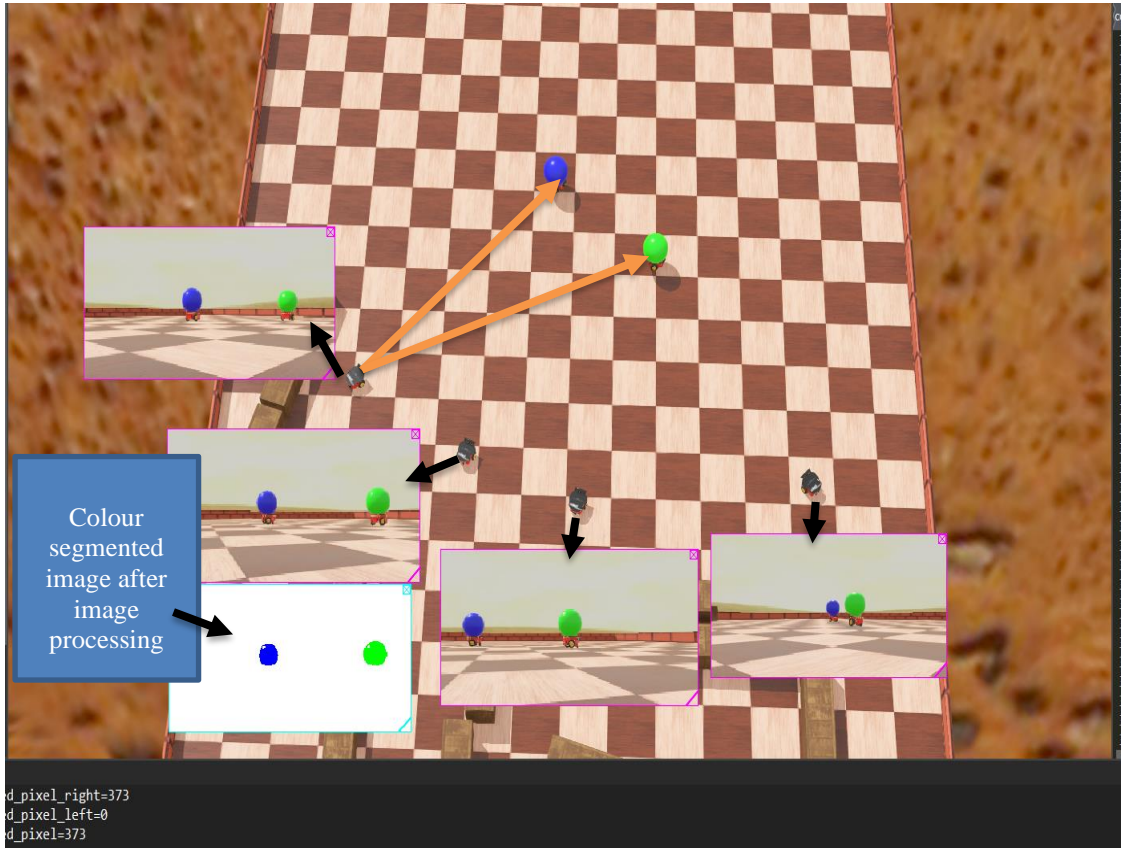

**Figure 27: Color segmented image for Robot 2**

Now as robots are able to detect these intruders, they also measure the distance from each intruder by using range finder sensor. The strategy is very simple that robot measures the distance of only those pixel which are green or blue and this way robot able to decide which robot need to follow.

During programming the maximum range which can be measured using range finder is set to 10 meters. Now let's consider the scenario shown in Figure 27. The measured distance values by each robot to both intruders are shown in Table 2. Now in this particular case Robot 1 follows blue intruder while Robot 2, Robot 3 and Robot 4 follows green intruder.

**Table 2: Distance of robots from intruders**

| Robots in Simulation | Distance from blue intruder (m) | Distance from green intruder (m) |
|---|---|---|
| Robot 1 | 6.05 | 6.55 |
| Robot 2 | 6.06 | 5.37 |
| Robot 3 | 6.28 | 5.27 |
| Robot 4 | 8.52 | 5.51 |

Now let's consider another case in which intruders position is changed as shown in Figure 28 and measure the distance values again which is mentioned in Table 3. The grey window is actually field of view of range finder which helps it to measure distances. So, Robot 1 and Robot 2 will follow green robot as it is closer to them and on the other hand Robot 3 and Robot 4 will follow blue robot.

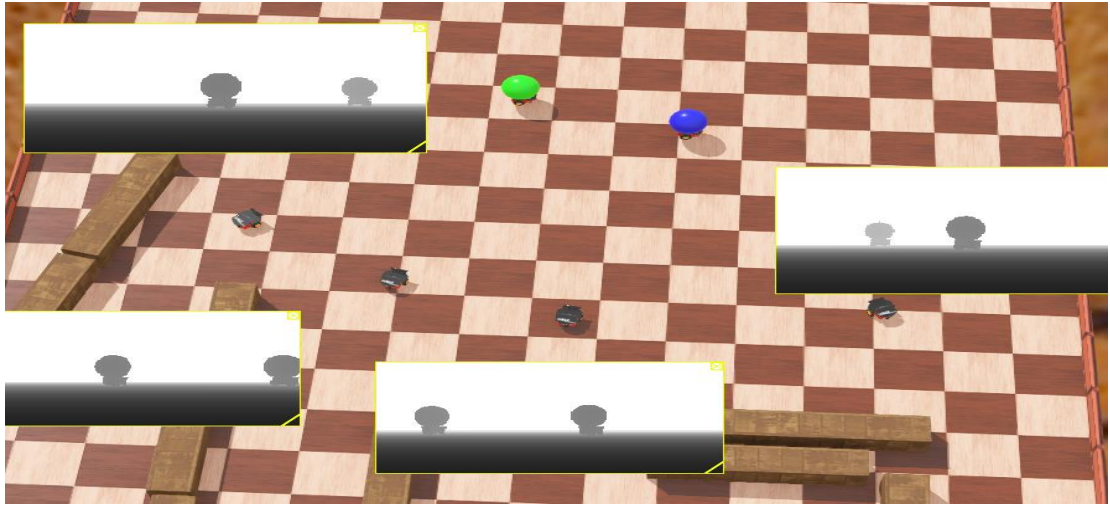

**Figure 28: Simulation environment in WEBOTS software**

**Table 3: Robots distances from intruders**

| Robots in Simulation | Distance from blue intruder (m) | Distance from green intruder (m) |
|---|---|---|
| Robot 1 | 6.35 | 5.05 |
| Robot 2 | 5.09 | 5.05 |
| Robot 3 | 4.99 | 5.44 |
| Robot 3 | 7.29 | 5.25 |

Now in below Figure 29 shows how robots come near the intruder and stop maintaining distance. Now when intruders are moved manually the robot follows them. When an object is further away from camera then it will appear smaller on camera field of view and on the hand when object comes closer than its size increases on camera field of view. So object size is actually depends on the number of pixels occupied in the camera field of view. So logic which is used to maintain a distance from intruder is dependent on number of pixels occupied by the intruder. If number of occupied pixels is greater than 3000 then robot will stop. But as the intruder is moving and number of pixels never increases so the robot also keep following intruder.

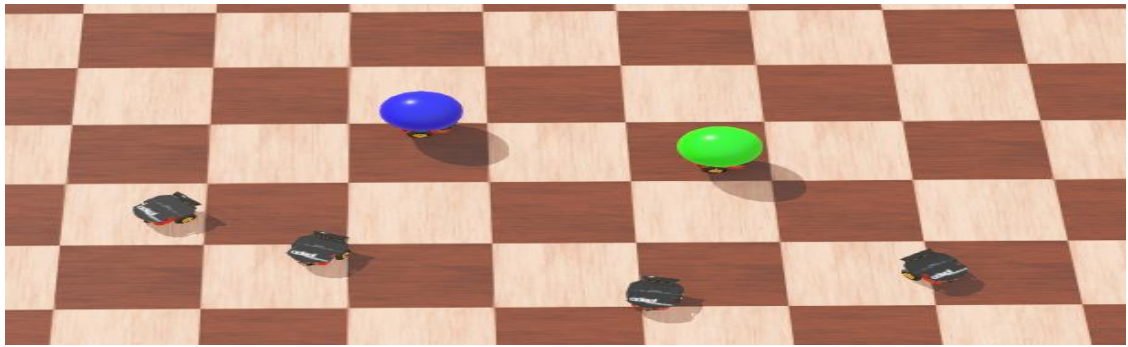

**Figure 29: Robots following intruders**

Now let's move green intruder robot and check the how friendly Robot 3 and Robot 4 follows the intruder robot. In simulation, green intruder moved in a straight line for 13 seconds and total distance covered by intruder is recorded as shown in Figure 30. Also at four different intervals of time, the distance of Robot 3 and Robot 4 is noted from green intruder as shown in Table 4.

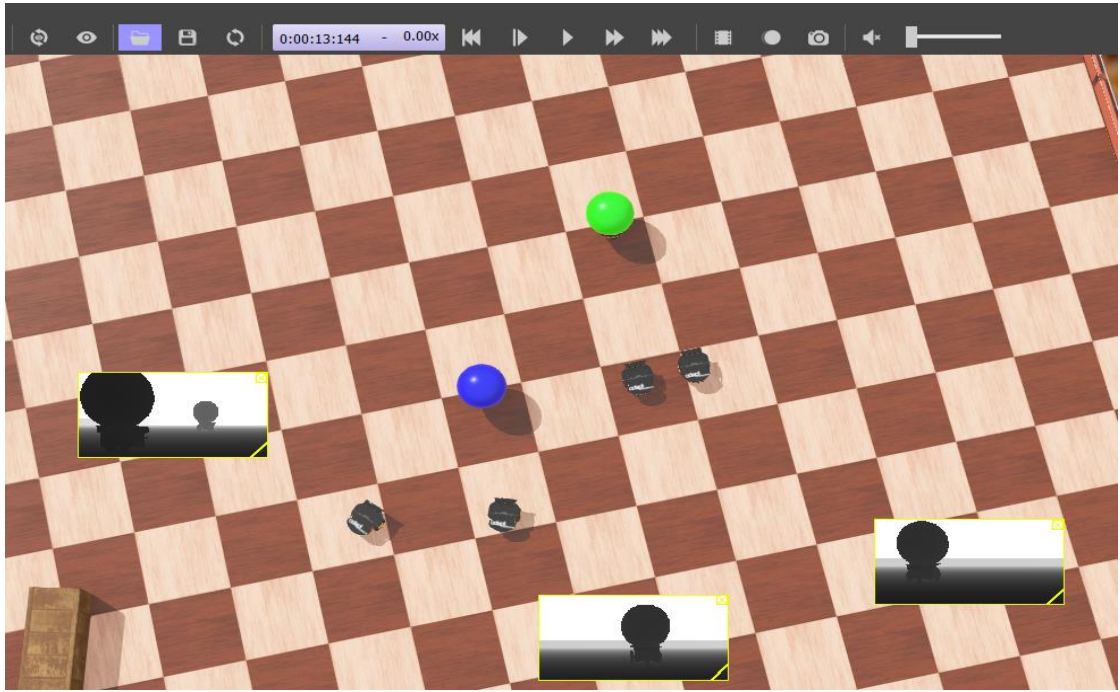

**Figure 30: Robot 3 and Robot 4 following green intruder**

**Table 4: Robot following an intruder**

| Time in sec (s) | Distance moved by Intruder robot (m) | Distance between Intruder and Robot 3 (m) | Distance between Intruder and Robot 4 (m) |
|---|---|---|---|
| 1 | 0 | 1.8 | 1.82 |
| 5 | 0.65 | 1.83 | 1.84 |
| 9 | 1.05 | 1.8 | 1.79 |
| 13 | 1.45 | 1.81 | 1.8 |

If these results are shown in graphical form then they can be seen in Figure 31.

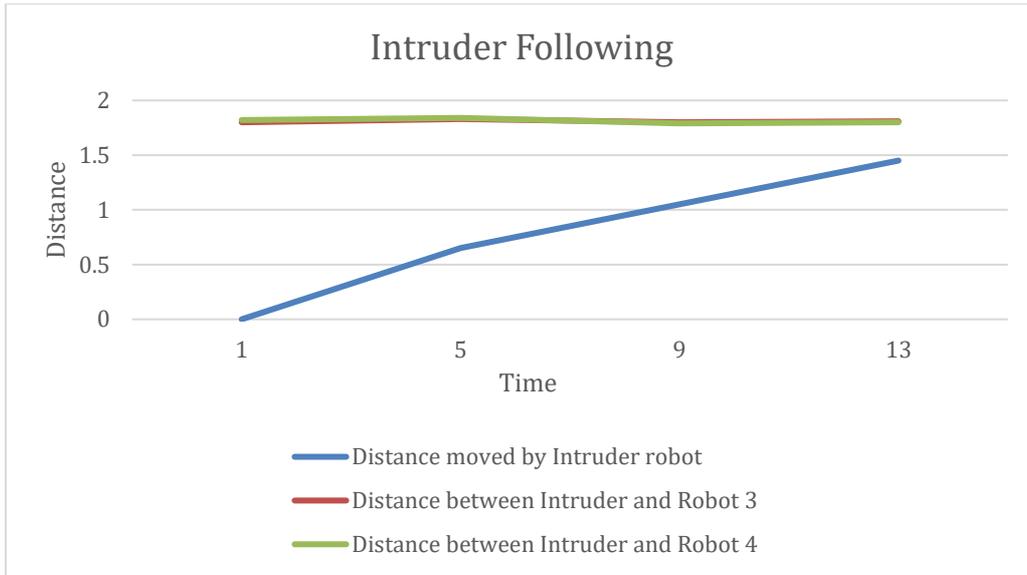

**Figure 31: Graph for Robot 3 and Robot 4 following Green intruder**

Now it can easily be deduced from above graph that even when intruder is moving from its initial position and covering distance as displayed by blue line in graph, the distance between intruder and both Robot 3 and Robot 4 remains almost same as represented by green and red straight lines. So those robots are following the intruder wherever it is moving by maintaining a minimum distance to avoid collision. The same strategy can be applied on other two robots and blue intruder and same results will be received. Hence, the aim of friendly robots following intruders in the environment is successfully achieved.

**4.4 Conclusion**

In this chapter, designing of simulation environment is discussed in detail. The methodology which is explained in Chapter three is used to perform all the simulation. Method of obtaining all the data from WEBOTS software simulation and using the data to show the results in the form of figures and tables is also explained. In the next Chapter six, the discussion of this obtained data in the light of aims and objectives of this report will be presented.

# CHAPTER 5.     DISCUSSION AND CONCLUSION

The results obtained above clearly shows the effectiveness of proposed strategies for this report project. There are multiple aims and objectives mentioned in Chapter one which are in consideration while getting these results. Here is the comparison of obtained results and objectives for this project

- The first objective is to develop a strategy for decentralized movement of multiple robots in a cluttered environment while avoiding obstacle. An arithmetic mean based navigation algorithm for robots to move between obstacles is developed and target of moving the robots in the middle of obstacles is successfully achieved. This navigation strategy is very novel in its implementation as compare to the navigation strategies of bug algorithm used by [7] and [22], of A* algorithm used by [10] and [17], of fuzzy logic controller algorithm used by [18], [19] and [20] and of hybrid algorithm used by [21].The results which are represented in Graph 1, Graph 2, Graph 3 and Graph 4 do authenticate that robots tend to stay in the middle of obstacles during their movement. The slight error between suggested path and path taken by robot is due to continuous change of distance between obstacles and robot needs to little time to settle on continuous changing path.
- The second objective is to construct a method for detecting intruders whose positions are unknown. For getting the desired results, two intruders having different colors are used. The simple image processing color segmentation technique using camera is implemented to detect intruders just like [14], [15] and [22]. The result shown in Figure 17 does indicate that all robots are able to detect intruders with proposed strategy. Also using color segmentation technique and showing these results in a separate display window in WEBOTS software is also successfully implemented, it helps in showing the desired results clearly to mention them in report report. Any advanced image processing technique like SIFT [33], SURF [34] or template matching [32] which can increase robot

- processing time is not used as aim is to just detect intruder robots and that is achieved by simple color segmentation technique.
- The third aim is to measure the distance between intruders and friendly robots and develop a mechanism to follow intruders. A range finder just like [5],[13], [16], [29] and [30] is used for this purpose and results are shown in Table 2 and Table 3. The distance is measured continuously when intruder is in sight of any friendly robot. Then intruder which is closer to the robot is selected as a primary target to be followed by using probability of danger algorithm. This aim is also been successfully achieved and results are displayed in Table 4. A graphical interpretation of results is also presented in Graph 5. This probability of danger algorithm for following multiple intruders is also novel in its implementation. The additional ability of a robot to change primary target when following an intruder and another intruder detected which is more closer than before is not part of additional hypothesis. But this additional functionality has also been achieved during simulation.
- In the last, successful integration of Kinect sensor's camera and range finder for intruder detection and following, while using sonar sensors for robot navigation is successfully implemented.

There are also some research questions defined in Chapter one which were focal point while finding results of this report. Here is the short description those challenges in light of results

- So Pioneer 3-DX is selected as a test robot which provides all the functionalities and control system for integrating all the sensors and devices together. This robot is readily available in software environment. All the results are successfully achieved by using this robot.
- For navigation of robot through obstacles, an arithmetic mean based strategy is successfully designed and implemented. And effectiveness of this strategy is proved via results.

- A camera is used for detection of different color intruder robots and results proved its efficacy. A probability of danger algorithm is designed and successfully implemented to follow intruders.

This navigation strategy works perfectly when robot is moving through an area having obstacles on both side of robot but issues may arise but not confirmed if an area having obstacles at different places of environment. The robot is going to avoid obstacles in this case too but detection of intruders strategy would be affected if intruder is not in sight of friendly robot.

Only intruders present in friendly robot field of view are considered at specific time and probability of danger algorithm is only apply to those intruders to make the robot to follow them follow them. The simulation environment is designed in such a way that all robots detect all intruders and make decision to follow them. Otherwise in different scenario if there is only one intruder in field of view of all robots then only that intruder will be followed.

Hence, the research questions, aims and objectives which are highlighted in this report are successfully explained and proved via results. But there are still a lot of improvement which can be achieved as explained in future recommendations.

## 5.1   Future Recommendations

For future work, performance of different algorithms with the performance of purposed algorithm can be compared to evaluate the best navigation strategy through obstacles. In this report, color segmentation technique is implemented but in future different image processing techniques can be integrated to this project to increase the efficiency of detecting objects even during the night time. Instead of two intruders more intruders can be introduced and a mechanism of one robot following one intruder randomly will also be step forward in this report. All the obstacles were considered static during the simulation of this project, so avoiding dynamic obstacles with dynamic intruders will have a great practical and industrial impact on the commercialization of ground robots.